%% file: CameraReady.tex
\documentclass[10pt,twocolumn,letterpaper]{article}

\usepackage{cvpr}              

\usepackage[title]{appendix}
\usepackage{titling}
\usepackage{graphicx}
\usepackage{amsmath}
\usepackage{amssymb}
\usepackage{booktabs}
\usepackage{color}
\usepackage{bm}
\usepackage{mathrsfs}
\usepackage{multirow,xspace}
\usepackage{makecell}
\usepackage{eucal}
\usepackage{flushend,cuted}
\usepackage{placeins}
\usepackage{bbding}
\usepackage{multirow}
\usepackage[ruled,vlined]{algorithm2e}
\usepackage{threeparttable}
\usepackage[pagebackref,breaklinks,colorlinks]{hyperref}

\usepackage[capitalize]{cleveref}
\crefname{section}{Sec.}{Secs.}
\Crefname{section}{Section}{Sections}
\Crefname{table}{Table}{Tables}
\crefname{table}{Tab.}{Tabs.}

\def\cvprPaperID{2507} 
\def\confName{CVPR}
\def\confYear{2022}

\begin{document}

\title{Attribute Surrogates Learning and Spectral Tokens Pooling \\
in Transformers for Few-shot Learning}


\author{{Yangji He$^{1,2}$ \qquad Weihan Liang$^{1,2}$ \qquad Dongyang Zhao$^{1,2}$ \qquad Hong-Yu Zhou$^3$ }
\\{Weifeng Ge$^{1,2}$\thanks{Corresponding author:wfge@fudan.edu.cn} \qquad Yizhou Yu $^3$ \qquad Wenqiang Zhang$^2$ } \\$^1$Nebula AI Group, School of Computer Science, Fudan University  \\ $^2$Shanghai Key Lab of Intelligent Information Processing\\$^3$Department of Computer Science, The University of Hong Kong\\    
}

\maketitle

\begin{abstract}
This paper presents new hierarchically cascaded transformers that can improve data efficiency through attribute surrogates learning and spectral tokens pooling. Vision transformers have recently been thought of as a promising alternative to convolutional neural networks for visual recognition. But when there is no sufficient data, it gets stuck in overfitting and shows inferior performance. To improve data efficiency, we propose hierarchically cascaded transformers that exploit intrinsic image structures through spectral tokens pooling and optimize the learnable parameters through latent attribute surrogates. The intrinsic image structure is utilized to reduce the ambiguity between foreground content and background noise by spectral tokens pooling. And the attribute surrogate learning scheme is designed to benefit from the rich visual information in image-label pairs instead of simple visual concepts assigned by their labels. Our Hierarchically Cascaded Transformers, called HCTransformers, is built upon a self-supervised learning framework DINO and is tested on several popular few-shot learning benchmarks.  

In the inductive setting, HCTransformers surpass the DINO baseline by a large margin of 9.7\% 5-way 1-shot accuracy and 9.17\% 5-way 5-shot accuracy on mini{\em ImageNet}, which demonstrates HCTransformers are efficient to extract discriminative features. Also, HCTransformers show clear advantages over SOTA few-shot classification methods in both 5-way 1-shot and 5-way 5-shot settings on four popular benchmark datasets, including mini{\em ImageNet}, tiered{\em ImageNet}, FC100, and CIFAR-FS. The trained weights and codes are available at \url{https://github.com/StomachCold/HCTransformers}.
\end{abstract}

\section{Introduction}
\label{sec:intro}

Vision transformers have recently become a prominent alternative to convolutional neural networks (CNNs) in many computer vision tasks, including image classification~\cite{dosovitskiy2020image,liu2021swin,srinivas2021bottleneck}, object detection~\cite{carion2020end,dai2021dynamic,zhu2021deformable,DBLP:conf/cvpr/DaiCLC21,sun2021rethinking}, semantic segmentation~\cite{xie2021segformer,Strudel_2021_ICCV,DBLP:conf/cvpr/ZhengLZZLWFFXT021,dong2021solq,guo2021sotr} and etc. However, according to~\cite{dosovitskiy2020image,liu2021swin}, transformers will work in low efficiency on a learning task with only few annotated samples. In this paper, we investigate whether vision transformers can still work well when the amount of labeled training data is extremely limited.

Few-shot learning~\cite{miniimagenet,Gidaris_2019_ICCV,lu2020learning} refers to the problem of learning from a very small amount of labeled data, which is expected to reduce the labeling cost, achieve a low-cost and quick model deployment, and shrink the gap between human intelligence and machine models. The key problem of few-shot learning is how to efficiently learn from the rich information hidden in annotated data. Inspired by the part-whole hierarchical concepts used in GraphFPN~\cite{zhao2021graphfpn} and GLOM~\cite{hinton2021represent}, part layout information of objects/scenes contains various visual information. If it can be embedded in vision transformers to guide the feature learning, we will get discriminative feature representations. Meanwhile, to avoid the concentration of visual information on single concepts, we need to expand the hidden information of image labels into a much more general semantic representation. Then how to mine such latent information and generate a complete description of visual concepts becomes important.

In this paper, we aim to improve the data efficiency in ViT~\cite{dosovitskiy2020image} for few-shot image classification. To be specific, we design a meta feature extractor composed of three consecutively cascaded transformers, each of which models the dependency of image regions at different semantic levels. The output tokens of a previous transformer are passed to a spectral tokens pooling layer to produce the input tokens for the subsequent. The spectral tokens pooling is partly based on spectral clustering~\cite{ng2002spectral,yan2009fast}, where features of tokens within same clusters are averaged to generate new token descriptors for the subsequent transformer. 
The motivation behind the spectral tokens pooling is to bring the image segmentation hierarchy into transformers. That means when the transformer performs self-attention, it needs to consider the image layout not simply through the positional embedding, but from the semantic relationship of different image regions. In our implementation, each token can be thought to represent some specific region within an image. We treat every token as a vertex in a graph and the token similarity matrix describes the edge connectivity. Thus, the spectral tokens pooling becomes an image segmentation problem and can be solved efficiently as that in normalized cut~\cite{shi2000normalized}.

In practice, we insert two spectral tokens pooling layers between three transformers. Since they capture the semantic dependencies of tokens in different hierarchies, we call them Hierarchically Cascaded Transformers (written as HCTransformers). Besides, we don't utilize the supervision information directly as that in other state-of-the-art few-shot learning methods~\cite{S2M2,IE,BML,TPMN}. 
Instead, we introduce a latent attribute surrogates learning scheme to learn robust representations of visual concepts. We hallucinate some latent semantic surrogates for each class to guide the learning of deep models. The latent semantic surrogates also have learnable parameters that can be jointly learnt with the parameters of transformers end-to-end. In fact,  it's a kind of weakly supervised learning by generalizing image-level annotation into attribute-level supervision. Based on such a latent attribute surrogate learning scheme, we avoid directly mapping an image into a single visual concept from a predefined set of object categories. 

The contributions of this paper are as follows:
\begin{itemize}
\item We employ ViT as the meta feature extractor for few-shot learning and propose Hierarchically Cascaded Transformers (HCTransformers), which greatly improve the data efficiency through attribute surrogates learning and spectral tokens pooling.

\item We propose a latent supervision propagation scheme for transformers in a weakly supervised manner. It converts the image label prediction task into a latent attribute surrogates learning problem. In this way, both the class and patch tokens can be supervised efficiently.

\item We introduce a novel spectral tokens pooling scheme to transformers. It models the dependency relationship of image regions in both the spatial layout and the semantic relationship. Due to such a mechanism, ViT can learn much more discriminative features at different semantic hierarchies.

\item Experiments demonstrate that our HCTransformers surpass its DINO baseline on {\em mini}ImageNet~\cite{miniimagenet}, and outperform other state-of-the-art algorithms significantly on multiple few-shot learning benchmarks, including {\em mini}ImageNet~\cite{miniimagenet}, {\em tiered}ImageNet~\cite{tieredimagenet}, and CIFAR-FS~\cite{CIFAR-FS} and FC-100~\cite{FC100}.
\end{itemize}

\section{Related Work}
\noindent\textbf{Meta-/Few-shot Learning.} Meta-learning or {\em``learning to learn''}~\cite{pfahringer2000meta,andrychowicz2016learning} refers to improving a learning task by learning over multiple learning episodes. Meta-learning has become the dominant paradigm for few-shot learning~\cite{finn2017model,sung2018learning}. Various meta-learning based methods have been proposed for few-shot image classification, such as MAML~\cite{finn2017model}, REGRAB~\cite{qu2020few}, TAML~\cite{jamal2019task}, MetaOptNet~\cite{lee2019meta}, and etc. However, according to \cite{chen2019closerfewshot,gidaris2018dynamic,qiao2018few}, training CNNs from scratch with meta-learning shows inferior performance in comparison to fine-tuning a CNN feature extractor pre-trained in a standard manner. There are also other methods focusing on better feature extraction~\cite{wu2019parn}, additional knowledge~\cite{wertheimer2019few}, knowledge transfer~\cite{li2019large}, and graph neural networks~\cite{kim2019edge}. Different from previous meta-learning methods, we introduce the inherent semantic hierarchies of images into transformers, and supervise the parameter learning with latent attribute surrogates. By this way, we alleviate the overfitting problem and get impressive results.\\



\noindent\textbf{Tokens Pooling in Transformers.} ViT \cite{dosovitskiy2020image} directly applies transformer architecture into vision tasks by splitting input image into $16 \times 16$ tokens via patch embedding. Despite impressive results on several vision benchmarks, vanilla transformer architectures focus on attending global information while neglecting local connections, which hinders the use of fine-grained image features thus leading to their data-hungry nature. Many subsequent works address this issue by establishing a progressive shrinking pyramid that allows models to explicitly process low-level patterns. There is a group of approaches that merge tokens within each fixed window into one to reduce the number of tokens~\cite{liu2021swin,wang2021pyramid, Wu_2021_ICCV, Heo_2021_ICCV, Yuan_2021_ICCV, Pan_2021_ICCV,chen2021psvit}. In contrast, the second group of methods drops this constraint and introduces more flexible selection scheme~\cite{rao2021dynamicvit,chen2021chasing,Yue_2021_ICCV,marin2021token}. While our HCTransformers  allow tokens to be adaptively merged with their neighboring tokens according to their spatial layout and semantic similarities.\\

\noindent\textbf{Supervise Patch Tokens in Transformers.} ViT\cite{dosovitskiy2020image} adds a $[cls]$ token to globally summarize the integral information of patch tokens and only this token directly receives supervision signals. However, other tokens maintain the ability to express distinctive patterns and may delicately assist final prediction. Some works proposed to remove the $[cls]$ token and construct a global token by integrating patch tokens via certain average pooling operation~\cite{liu2021swin, Pan_2021_ICCV,rao2021global,chen2021psvit}. LV-ViT~\cite{jiang2021all} explored the possibility to jointly utilize $[cls]$ token and patch tokens. It reformulates the classification task with the token labeling problem. Likewise, So-ViT \cite{SoViT} applied a second-order and cross-covariance pooling to visual tokens, which is combined with the $[cls]$ token for final classification. Our method shares similar intuition with these two methods, but the difference is apparent. We integrate patch tokens as a weighted sum where scores are calculated based on their connections with the global $[cls]$ token, which aims to mostly utilize significant patch tokens. Besides, we suppose the integrated patch tokens do not share the feature space with the $[cls]$ token and supervise them within their own feature space.

\begin{figure*}[htbp]  
\centering  
\includegraphics[width=1\textwidth]{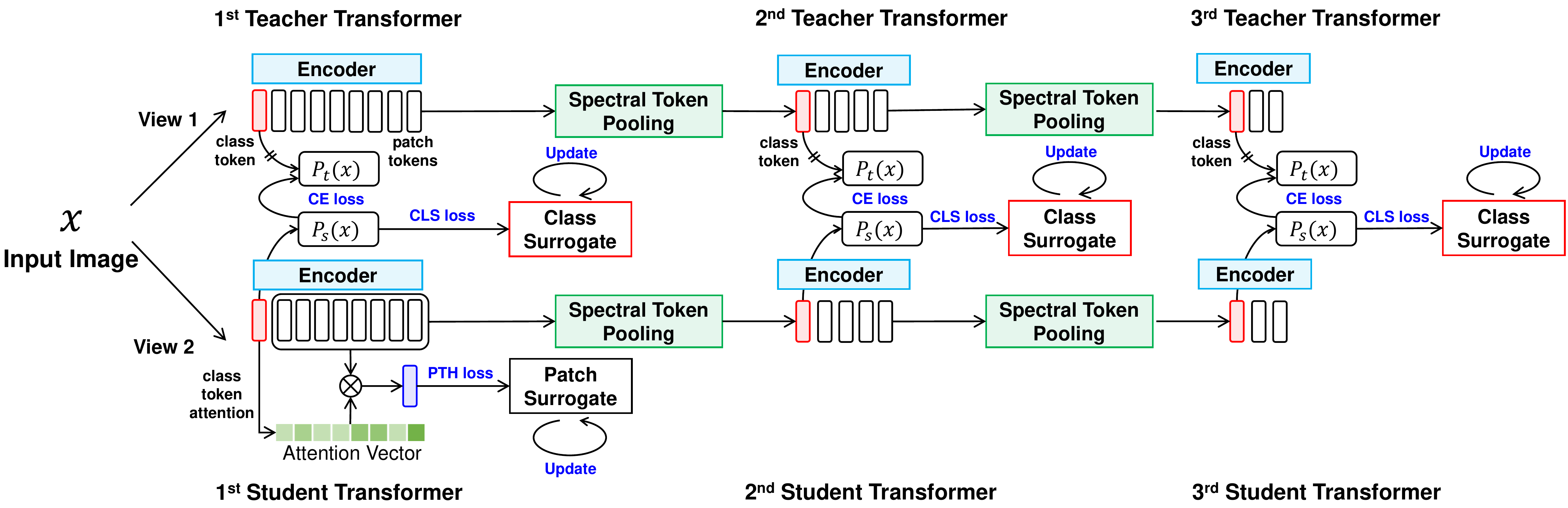}  
\caption{Illustration of the overall structure. HCTransformers contain three sets of cascaded transformer networks, each corresponding to a teacher-student knowledge distillation framework. During training, every student transformer builds and updates a surrogate descriptor for each category. The patch tokens integrated through the attention map are also used to generate patch surrogate descriptors in the first stage. In between every two transformer sets, a spectral tokens pooling layer is used to down-sample the patch token number by $\frac{1}{2}$ for information aggregation.}  
\label{fig:network}
\end{figure*}

\section{Hierarchically Cascaded Transformers}
\subsection{Overview}
The core of HCTransformers is to make the full usage of the annotated data to train transformers as a strong meta feature extractor for few-shot learning. To benefit from existing self-supervised learning techniques~\cite{grill2020bootstrap,caron2021emerging}, we use DINO~\cite{caron2021emerging} as our base learning framework and multi-crop strategy~\cite{caron2020unsupervised} to conduct knowledge distillation~\cite{hinton2015distilling}. In the pretraining stage of few-shot learning, we can access image labels. We design a latent attribute surrogates learning scheme for both patch tokens and the $[cls]$ tokens to avoid directly learning from labels. To incorporate the semantic hierarchy into transformers, we insert a spectral tokens pooling layer between two ViTs~\cite{dosovitskiy2020image}. The output similarity matrix of patches can be used to conduct spectral clustering to segment patches into disjoint regions. Then the feature average of patches in the same region is treated as a new token feature, which captures higher-level semantic information. In \cref{fig:network}, we illustrate the complete pipeline of our proposed method. \\

\subsection{Preliminary}
Similar to BYOL~\cite{grill2020bootstrap}, DINO~\cite{caron2021emerging} employs a knowledge distillation framework for self-supervised learning with two homogeneous networks: teacher and student, where the teacher's parameter ($\theta_t$) is an exponential moving average results of the updated student's parameter ($\theta_s$) on the foundation of ViT~\cite{dosovitskiy2020image} architecture. Each network consists of a transformer encoder and a classification head (i.e., a multi-layer perceptron and one fully connected layer).

Given an input image $x \in \mathbb{R}^{H\times W\times C}$, DINO first reshapes $x$ into a sequence of flattened 2D patch tokens $t_p \in \mathbb{R}^{N \times D}$. $t_p$ is then concatenated with a learnable class token $t_{c} \in \mathbb{R}^{1 \times D}$ for an augmented sequence $T\in \mathbb{R}^{(N+1)\times D}$. Here, $\{H, W\}$ denote the spatial resolution of the input image, $C$ stands for the image channel number. $N = HW/P^2$ is the resulting number of patches, where $P$ denotes the patch size. $D$ represents the encoded feature dimension. After passing through the transformer encoder, the sequence enhances its representation through self-attention. We denote the enhanced patch tokens as $f_p (x) \in \mathbb{R}^{N\times D}$. Likewise, the enhanced $[cls]$ token is denoted as $f_c (x)\in \mathbb{R}^{1\times D}$. Moreover, the token similarity matrix ${\bf A} \in \mathbb{R}^{(N+1) \times (N+1)}$ can be acquired from the self-attention process. Afterwards, only the encoded $[cls]$ token is used for final prediction. We pass this $[cls]$ token to the projection head to map it into a higher $D^{\prime}$-dimension, which is denoted as $P(x)$. 
 
Besides, DINO employs a multi-crop training augmentation scheme. For any given image $x$, it constructs a set $V$ of the subregions, including two global views, $x_1^g$ and $x_2^g$ and $m$ local views. DINO minimizes the following loss to encourage the ``local-to-global'' prediction consistency:

\begin{equation}
\mathcal{L}_{\rm DINO} = \frac{1}{M} \sum\limits_{x_g \in\left \{ x_1^g,x_2^g \right \}}\sum\limits_{
		x^{\prime} \in V,\ x^{\prime} \neq x_g}- P_t(x_g)\ {\rm \log}{P_s(x^{\prime})},
\label{eq:dino_loss}
\end{equation}
where $M$ is the pair number $2\times (m+1)$. $P_t(x_g)$, $P_s(x^{\prime})$ are the outputs of teacher and student networks, respectively.

\subsection{Jointly Attribute Surrogates and Parameters Learning}
\label{sec:feature}
We design a latent supervision propagation scheme for transformers to avoid supervising the parameter learning only through a quite limited amount of one-hot labels. For each visual concept $y$ in the label space, we aim to learn a semantic attribute surrogate $z(y) \in \mathbb{R}^{1\times D^{\prime}}$ for it,
\begin{equation}
y \rightarrow z(y),
\label{eq:surrogates}
\end{equation}
 where $D^{\prime}$ is the surrogate descriptor dimension. When there are $C$ classes, the surrogate descriptors ${\bf Z} \in \mathbb{R}^{C \times D^{\prime}}$ would contain $C$ entries. During the learning process, the supervision is passed through surrogates to supervise the student's parameter learning. At the same time, these surrogates need to be learnt at the same time. Supposing the supervised learning objective of the student is $\mathcal{L}_{\rm surr}$, then the parameters $\theta_s$ of the student and its associated surrogates can be updated with the following equations
 \begin{equation}
\theta_s^{t+1} = \theta_s^{t}- \gamma_1   \frac{\partial \mathcal{L}_{\rm surr}}{\partial \theta_s^{t}} ,
\label{eq:parameter_loss}
\end{equation}

 \begin{equation}
z(y)^{t+1} = z(y)^{t} - \gamma_2   \frac{\partial \mathcal{L}_{\rm surr}}{\partial z(y)^{t}},
\label{eq:surrogates_loss}
\end{equation} 
where $\gamma_1$ and $\gamma_2$ are the learning rates. During the initialization, both $\theta_s$ and $\bf Z$ are initialized with Gaussian noises. Following the settings in DINO ~\cite{caron2021emerging}, we use the AdamW optimizer~\cite{loshchilov2018fixing} with momentum to update both $\theta_s$ and ${\bf Z}$ with linear scaling rule~\cite{goyal2017accurate}, which is a slightly different with that in the center loss~\cite{wen2016discriminative}.

To make the fully advantage of transformers, we learn semantic attribute surrogates for both patch and class tokens respectively.\\


\noindent\textbf{Supervise the Class Token.} In DINO~\cite{caron2021emerging} and other knowledge distillation methods, the student network produces probability over $D^{\prime}$ dimensions. Different from traditional supervised learning paradigms, $D^{\prime}$ is set to a quite large number without considering the dataset's real class number. In this paper, we set $D^{\prime} = 8192$. To be consistent with the teacher-student knowledge distillation in DINO, we use the surrogate loss to supervise the probability distribution learning for every class. Then the surrogate descriptor ${ z_c}(y) \in \mathbb{R}^{8192}$ of the class $y$ is a vector on 8192 dimensions. We normalize ${ z_c}(y)$ with the Softmax operation to get an attribute distribution $\overline{{ z_c}}(y)$. Following the annotations in Eq.~\ref{eq:dino_loss}, the class token loss becomes:
\begin{equation}
		\mathcal{L}^{\rm cls}_{\rm surr} = \frac{1}{2}  \sum\limits_{x_g \in\left \{ x_1^g,x_2^g \right \}} D_{\rm KL}(P_s(x_g)||\overline{{ z_c}}(y)),
	\label{eq:cls_loss}
\end{equation}
where $y$ is the label of the input image $x$, and $D_{\rm KL}$ is the Kullback–Leibler divergence. Note that only global views are involved here considering that local views may introduce negative effects when updating class centers due to the loss of information.\\


\noindent\textbf{Supervise Patch Tokens.} 
In transformers, patch tokens are hard to be supervised due to the lack of patch-level annotations. To supervise patch tokens, we firstly aggregate the patch token features $f_p(x)$ to generate a global descriptor of an image $x$ by applying the attention map ${\bf A}_c (x) \in \mathbb{R}^{1 \times N}$:
\begin{equation}
	F_p(x) = {\bf A}_c (x) f_p (x)
	\label{eq:patch_aggre}
\end{equation}
where ${\bf A}_c (x)$ denotes the similarity matrix that can be acquired by calculating the similarity between the $[cls]$ token and each patch token. ${F}_p (x)\in \mathbb{R}^{1 \times D}$ is the patch token feature. We treat each entry in ${F}_p (x)$ as a description about some semantic attributes, and normalize it with a Softmax operation to convert it into an attribute distribution ${\overline{F}}_p (x) = {\rm Softmax}~ {F}_p (x)$. Similarly with that in the class token, we have an attribute surrogate $z_p\left ( y \right )$ for each class. The patch token loss becomes:
\begin{equation}
	\mathcal{L}_{\rm surr}^{\rm pth} = \frac{1}{2}  \sum\limits_{x_g \in\left \{ x_1^g,x_2^g \right \}} D_{\rm KL}({\overline{F}}_p (x)||\overline{z}_p\left ( y \right )),
	\label{eq:patch_loss}
\end{equation}
where $\overline{z}_p\left ( y \right ) = {\rm Softmax}~ {z}_p\left ( y \right )$. Only the global views are applied for same consideration as aforementioned.


\begin{figure}[t]  
\centering  
\includegraphics[width=1.0\columnwidth]{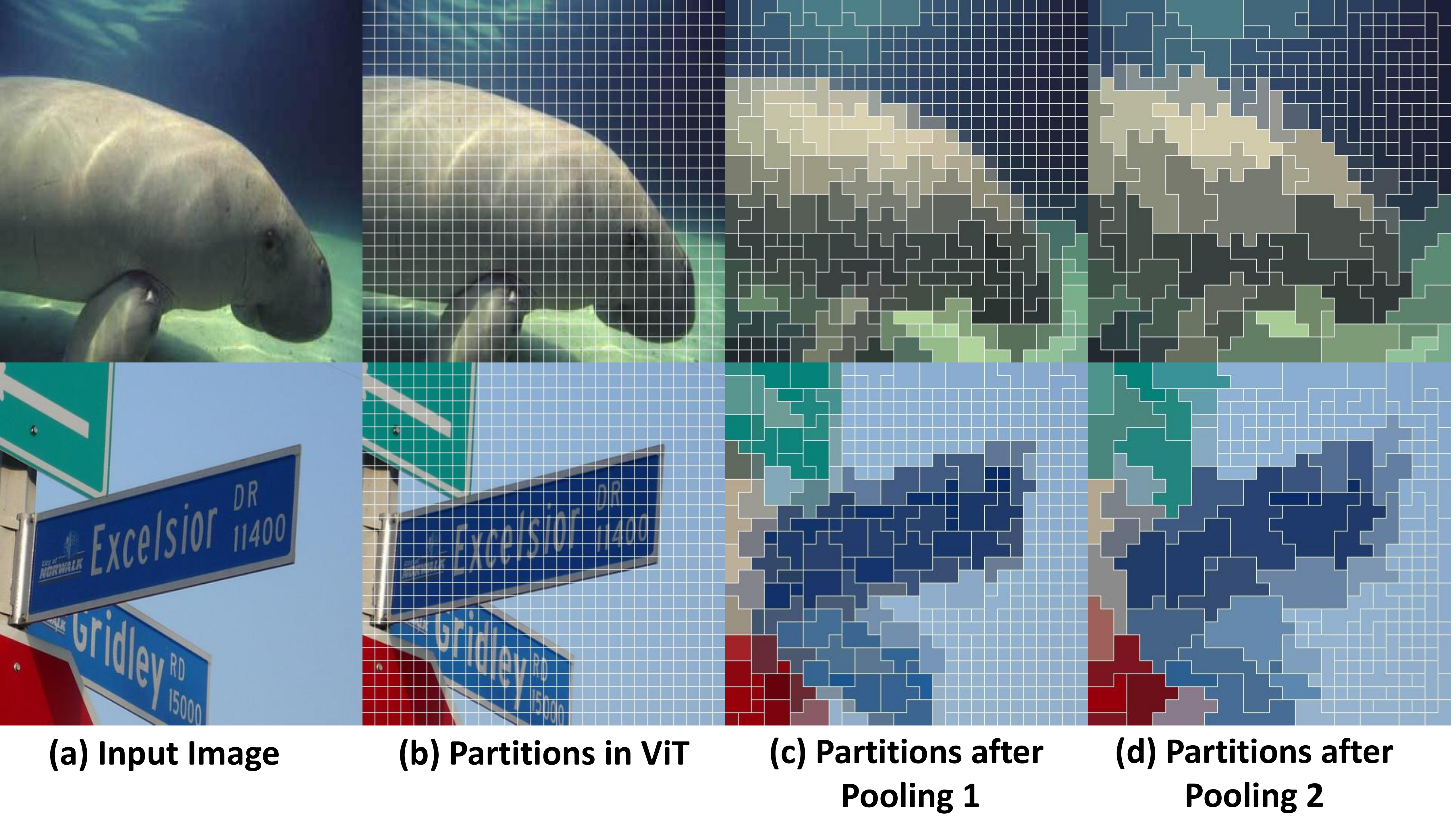}  
\caption{Visualized tokens pooling results. After spectral tokens pooling operations, adjacent tokens with similar semantics are clustered into one. (c) and (d) shows that our clustering results are well consistent with the image's basic structure. The pixel colors in the same cluster are averaged.}  
\label{fig:cluster}
\end{figure}
\subsection{Spectral Tokens Pooling}
\label{sec:pooling}
Multi-scale hierarchical structures have been proven effective in both convolutional neural networks~\cite{lin2017feature,jha2020doubleu} and transformers~\cite{liu2021swin}. We aim to improve the feature discriminative ability through embedding hierarchical structures into transformers. Different from the grid pooling scheme in Swin Transformer~\cite{liu2021swin}, here we exploit the irregular pooling method to match the image structures with more flexibility. Since transformers will generate self-attention matrices among tokens, it provides a strong prior for spectral clustering algorithms to segment tokens according to both their semantic similarities and the spatial layout. So we propose a spectral clustering-based pooling method, called spectral tokens pooling.

For the $N$ patch tokens in a ViT, we retrieve attention matrix between patches ${\bf A_p}$ $\in \mathbb{R}^{N \times N} $ from $\bf A$. To bring in spatial consistency, we maintain an adjacency matrix ${\bf H} \in \mathbb{R}^{N \times N}$ to reflect the neighborhood relationship. In our design, only 8-connected neighboring tokens are connected with the center token. We use the following formula to retrieve a symmetric $\bf S$ matrix. 
\begin{equation}
\bf S = \bf A_p \odot \bf H + \bf A_p^\mathsf{T} \odot \bf H^\mathsf{T}.
	\label{eq:patch_surrogate}
\end{equation}
Through the spatial constraint, the $\bf S$ matrix can be viewed as a sparse matrix for computation acceleration.
We then perform the Softmax operation on each row of $\bf S $ to get the final adjacency weight matrix $\bf S^{\prime}$. The spectral clustering algorithms~\cite{ng2002spectral,shi2000normalized,xu2009fast} are exploited to partition patch tokens into $N^{\prime}$ clusters $\mathcal{T^{prime}}=\left \{\mathcal{T}_1, \mathcal{T}_2,...,\mathcal{T}_{N^{\prime}} \right \}$ and generate new tokens with Algorithm~\ref{alg:stc}. During the backward stage, gradients of token clusters are copied to each of the averaged tokens. We implement the spectral tokens pooling with PyTorch. Fig. \ref{fig:cluster} visualizes results of two consecutive spectral tokens pooling.

\begin{algorithm}[]
	\caption{Spectral Tokens Pooling Algorithm}
	\label{alg:stc}
	\LinesNumbered
	\KwIn{Tokens $\bf T \in \mathbb{R}^{N \times D} $, adjacency weight matrix $\bf S^{\prime} \in \mathbb{R}^{N \times N}$, number $N^{\prime}$ clusters to construct.}
	\KwOut{Tokens $\bf T^{\prime} \in \mathbb{R}^{N^{\prime} \times D} $.}
	Compute the normalized Laplacian $\bf L$ from $\bf S ^{\prime}$;\\
	Compute the first $N^{\prime}$ eigenvectors $\mu_1,\mu_2,...,\mu_{N^{\prime}}$ of L;\\
	Construct a matrix $\bf U \in \mathbb{R}^{N \times N^{\prime}}$ to contain the vectors $\mu_1,\mu_2,...,\mu_{N^{\prime}}$ as columns;\\
	For $i=1, ..., n$, let $v_i \in \mathbb{R}^{N^{\prime}}$ be the vector corresponding to the $i$-th row of $U$;\\
	Cluster the points $(v_i)_{i=1,...,N}\in \mathbb{R}^{N^{\prime}}$ with K-means algorithm into clusters $\mathcal{T}=\left \{\mathcal{T}_1, \mathcal{T}_2,...,\mathcal{T}_{N^{\prime}} \right \}$;\\
	Features of tokens in the same clusters are averaged to generate new tokens  $\bf T^{\prime} \in \mathbb{R}^{N^{\prime} \times D} $.\\	
\end{algorithm}

\subsection{Training Strategy of HCTransformers}
In our design, two spectral tokens pooling layers are inserted into three different transformers. That means the outputs of previous transformers are sent to the subsequent transformers after performing pooling operations. In this way, tokens are organized with different semantic hierarchies. For different transformers, we set the output token numbers to 784, 392, 196, respectively. 

Since computing eigenvectors in spectral tokens pooling is time consuming ( 21.3 im/sec and 75.2 im/sec in two pooling layers respectively), we don't jointly all the three transformers end-to-end. The training is divided into two stages. In the first stage, we train the first two transformer as the same setting as DINO~\cite{caron2021emerging} with the loss function bellow, 
\begin{equation}
\mathcal{L}_{\rm stage1}= \mathcal{L}_{\rm DINO} + \alpha \mathcal{L}^{\rm cls}_{\rm surr} + \beta \mathcal{L}^{\rm pth}_{\rm surr}.
	\label{eq:final_1_loss}
\end{equation}
Then we frozen parameters of the first two transformer and train the subsequent two sets of transformers jointly with the same loss function in Eq.~\ref{eq:final_23_loss}. We only supervise on the $[cls]$ token for higher efficiency in the second stage. Since features produced by the first transformer already have strong discriminative ability, the training of subsequent transformers converges quickly in several epochs. 
\begin{equation}
\mathcal{L}_{\rm stage2}= \mathcal{L}_{\rm DINO} + \alpha \mathcal{L}^{\rm cls}_{\rm surr}
	\label{eq:final_23_loss}
\end{equation}
The weights $\alpha$ and $\beta$ are set to 1 and 0.1 in this work.


\begin{table*}[h]
\centering

  \small
  \begin{tabular}{c c cc cc}
    
    \toprule
    \multicolumn{1}{c}{\multirow{2}{*}{\textbf{Method}}}  & 
    \multicolumn{1}{c}{\multirow{2}{*}{\textbf{backbone}}} & \multicolumn{2}{c}{\textbf{\textit{mini}Imagenet}} & \multicolumn{2}{c}{\textbf{\textit{tiered}Imagenet}} \\
    \multicolumn{1}{c}{}   & \multicolumn{1}{c}{} & \textbf{1-shot}  & \multicolumn{1}{c}{\textbf{5-shot}} & \textbf{1-shot}  & \textbf{5-shot} \\
    \midrule

{DeepEMD}\cite{DeepEMD} &    \emph{ResNet-12} &   65.91  $\pm$ 0.82  &   82.41  $\pm$ 0.56  &   71.16  $\pm$ 0.87  &   86.03  $\pm$ 0.58 \\
{IE}\cite{IE} &    \emph{ResNet-12} &   67.28  $\pm$ 0.80  &   84.78  $\pm$ 0.52  &   72.21  $\pm$ 0.90  &   87.08  $\pm$ 0.58 \\
{DMF}\cite{DMF} &    \emph{ResNet-12} &   67.76  $\pm$ 0.46  &   82.71  $\pm$ 0.31  &    71.89  $\pm$ 0.52  &   85.96  $\pm$ 0.35 \\
{BML}\cite{BML} &    \emph{ResNet-12} &   67.04  $\pm$ 0.63  &   83.63  $\pm$ 0.29  &   68.99  $\pm$ 0.50  &   85.49  $\pm$ 0.34 \\
{PAL}\cite{PAL} &    \emph{ResNet-12} &  {\color{blue}69.37  $\pm$ 0.64} &   84.40  $\pm$ 0.44  &   72.25  $\pm$ 0.72  &   86.95  $\pm$ 0.47 \\
{METAQDA}\cite{METAQDA} &    \emph{WRN} &   67.38  $\pm$ 0.55  &   84.27  $\pm$ 0.75  &   74.29  $\pm$ 0.66  &   {\color{green}89.41  $\pm$ 0.77} \\
{TPMN}\cite{TPMN} &    \emph{ResNet-12} &   67.64  $\pm$ 0.63  &   83.44  $\pm$ 0.43  &   72.24  $\pm$ 0.70  &   86.55  $\pm$ 0.63 \\
{MN +  MC}\cite{MetaNavigator} &    \emph{ResNet-12} &   67.14  $\pm$ 0.80  &   83.82  $\pm$ 0.51  &   {\color{green}74.58  $\pm$ 0.88}  &   86.73  $\pm$ 0.61 \\
{DC}\cite{DC} &   \emph{WRN-28-10} &   68.57  $\pm$ 0.55  &   82.88  $\pm$ 0.42  &   {\color{blue}78.19  $\pm$ 0.25}  &   {\color{blue}89.90  $\pm$ 0.41} \\
{MELR}\cite{MELR} &   \emph{ResNet-12} &   67.40  $\pm$ 0.43  &   83.40  $\pm$ 0.28  &   72.14  $\pm$ 0.51  &   87.01  $\pm$ 0.35 \\
{COSOC}\cite{COSOC} &    \emph{ResNet-12} &   {\color{green}69.28  $\pm$ 0.49}  &  {\color{blue} 85.16  $\pm$ 0.42 } &   73.57  $\pm$ 0.43  &   87.57  $\pm$ 0.10 \\
{CSEI}\cite{CSEI} &    \emph{ResNet-12} &   68.94  $\pm$ 0.28  &   {\color{green}85.07  $\pm$ 0.50}  &   73.76  $\pm$ 0.32  &   87.83  $\pm$ 0.59 \\
{CNL}\cite{CNL} &    \emph{ResNet-12} &  67.96   $\pm$ 0.98 &  83.36  $\pm$ 0.51 &  73.42  $\pm$ 0.95 &  87.72  $\pm$ 0.75\\

\hline

\textbf{Baseline-Cosine}  &\emph{ViT-S}  & 52.92 $\pm$ 0.17 &  65.04 $\pm$ 0.14 & 66.04 $\pm$ 0.20  & 78.05 $\pm$ 0.16\\

\textbf{Ours-Cosine}  & \emph{ViT-S}  & {\color{red}74.74 $\pm$ 0.17} & 85.66  $\pm$ 0.10  & {\color{red}79.67 $\pm$ 0.20}  & 89.27  $\pm$ 0.13 \\ 

\textbf{Ours-Classifier}  & \emph{ViT-S}  & 74.62 $\pm$ 0.20 & {\color{red}89.19  $\pm$ 0.13} & 79.57  $\pm$ 0.20  & {\color{red}91.72  $\pm$ 0.11} \\ 

    \bottomrule
  \end{tabular}
 
  \caption{Comparison with the state-of-the-art 5-way 1-shot and 5-way 5-shot performance with 95\% confidence intervals on \textbf{\textit{mini}Imagenet} and \textbf{\textit{tiered}Imagenet}. \emph{ViT-S} is our baseline. Top three results are shown in {\color{red}red}, {\color{blue}blue} and {\color{green}green} based on their relative rankings.} 
  \label{tab:sota1}
\end{table*}

\section{Experiments}
\subsection{Datasets}
We perform experiments on four popular benckmark datasets for few-shot classification, including {\em mini}ImageNet~\cite{miniimagenet}, {\em tiered}ImageNet~\cite{tieredimagenet}, CIFAR-FS~\cite{CIFAR-FS}, and FC100~\cite{FC100}.
\textbf{{\em mini}ImageNet}~\cite{miniimagenet} contains 100 classes from the ImageNet~\cite{imagenet}~\cite{CIFAR-FS}, randomly split into 64 bases, 16 validation, and 20 novel classes, and each class contains 600 images. \textbf{{\em tiered}ImageNet}~\cite{tieredimagenet} contains 608 classes from 34 super-classes of the ImageNet, randomly split into 351 bases, 97 validation, and 160 novel classes. There are 779,165 images in total. \textbf{CIFAR-FS}~\cite{CIFAR-FS} contains 100 classes from the CIFAR100~\cite{CIFAR100}, randomly split into 64 bases, 16 validation, adn 20 novel calsses, and each class contains 600 images. \textbf{FC100}~\cite{FC100} contains 100 classes from 36 super-classes of the CIFAR100, where 36 super-classes were split into 12 base (including 60 classes), 4 validation (including 20 classes) and 4 novel (including 20 classes) super-classes, and each class contains 600 images. 

\subsection{Implementation Details}
All experiments are run on ViT-S/8 (8 is the size of each patch). We adopt the multi-crop strategy of DINO, we randomly crop and resize an image to 2 global images at resolution $224^2$ and 8 local images at resolution $96^2$. The teacher networks take in the 2 global views and the student networks take in all 10 image corps. In the first stage, to train a reliable meta feature extractor, we set $\alpha$ and $\beta$ to 1 and 0.1, respectively, then the gradients produced by $\mathcal{L}_{\rm DINO}$, $\mathcal{L}^{\rm cls}_{\rm surr}$ and $\mathcal{L}^{\rm pth}_{\rm surr}$ are in the same order of magnitude. Other specific parameters are inherited from DINO. In the second stage for spectral tokens pooling, we employ K-mean++ to conduct clustering. The $[cls]$ token input for the latter pooling transformer is initialized by the former's transformed $[cls]$ token for efficient training.

\noindent\textbf{Evaluation.} We evaluate experiments on 5-way 1-shot and 5-way 5-shot classification. For each task, we randomly select 5 categories. In each category, we use 1 or 5 labeled images as support data and another 599 or 595 unlabeled images of the same category as novel data. The reported results are the averaged classification accuracy over 10,000 tasks. During the meta test, we don't fuse the features of three student transformers. We actually use the validation set to select the class token features of the second student transformer to generate the final feature representation. For module ablations, we use the class token feature of individual transformer as the output. We use the simple Cosine classifier and the linear classifier in S2M2\cite{S2M2} to predict query labels.

\subsection{Comparisons with State-of-the-art Results}

\cref{tab:sota1} shows the 1-shot and 5-shot comparison results with the latest state-of-the-art (SOTA) methods on \emph{mini}Imagenet~\cite{miniimagenet} and \emph{tiered}Imagenet~\cite{tieredimagenet}. We outperform previous SOTA results by great margins with simple classifiers. For instance, on \emph{mini}Imagnet, HCTransformers surpasses SOTAs by 5.37\% (1-shot) and 4.03\% (5-shot), respectively. When we turn to \emph{tiered}Imagenet, our method outperforms the most recent DC~\cite{DC} by 1.48\% and 1.81\% on 1-shot and 5-shot, respectively. Compared with DC which borrows class statistics from the base training set, we don't need to do this and our classifier is lightweight. Another evidence is that the margin between our method and the third-best approach is 5.09\% on 1-shot, which helps validate our contribution. We give credit to our network structure for such impressive results, which can learn a lot of inherent information in data and maintain good generalization ability.

\cref{tab:sota2} and \cref{tab:sota3} display the results on small-resolution datasets, i.e., CIFAR-FS and FC100. HCTransformers show comparable or better results in these low-resolution settings: 1-shot ($1.02\%$) and 5-shot ($0.76\%$) on CIFAR-FS; 1-shot (0.51\%) and 5-shot (1.12\%) on FC100.  We observe that on the small resolution datasets, we don't surpass previous SOTA methods too much. We attribute this to the patching mechanism of ViT. When the image resolution is small, such as $32^2$, it is difficult to retrieve useful representation from cropped patches with limited numbers of real pixels. Similarly, DeepEMD~\cite{DeepEMD} also mentioned that patch cropping would have negative impacts on the small resolution images. However, our method still achieves the new SOTA results on both of these two benchmarks.

\begin{table}[h]
\centering
 \normalsize
 \scalebox{0.83}{
  \begin{tabular}{c c cc}
    
    \toprule
    \multicolumn{1}{c }{\multirow{2}{*}{\textbf{Method}}} & 
    \multicolumn{1}{c }{\multirow{2}{*}{\textbf{backbone}}} & \multicolumn{2}{c}{\textbf{CIFAR-FS}}  \\
    \multicolumn{1}{c }{}  &  \multicolumn{1}{c }{} & \multicolumn{1}{c}{\textbf{1-shot}}  & \multicolumn{1}{c}{\textbf{5-shot}}  \\
    \midrule
    {DSN-MR}\cite{DSN-MR}  &  \emph{ResNet-12} &   75.60  $\pm$ 0.90  &  86.20   $\pm$ 0.60\\
    {TPMN}\cite{TPMN}  &  \emph{ResNet-12} &   75.50 $\pm$ 0.90 &  87.20 $\pm$ 0.60\\
    {IE}\cite{IE}  &  \emph{ResNet-12} &   {\color{blue}77.87   $\pm$ 0.85}  &   {\color{blue}89.74 $\pm$ 0.57}\\
    {PSST}\cite{PSST}  &  \emph{WRN-28-10} &   77.02   $\pm$ 0.38  &  88.45    $\pm$ 0.35\\
    {BML}\cite{BML}  &  \emph{ResNet-12} &   73.45  $\pm$ 0.47  &  88.04   $\pm$ 0.33\\
    {PAL}\cite{PAL}  &  \emph{ResNet-12} &   {\color{green}77.10 $\pm$ 0.70} & 88.00 $\pm$ 0.50\\
    {MN +  MC}\cite{MetaNavigator}  &  \emph{ResNet-12} &   74.63 $\pm$ 0.91 &  86.45 $\pm$ 0.59\\
    {RENet}\cite{RENet}  &  \emph{ResNet-12} &   74.51 $\pm$ 0.46  &  86.60 $\pm$ 0.32\\
    {METAQDA}\cite{METAQDA}  &  \emph{WRN} &   75.95 $\pm$ 0.59  &  {\color{green}88.72 $\pm$ 0.79}\\
    {ConstellationNet}\cite{ConstellationNet}  &  \emph{ResNet-12} &   75.40   $\pm$ 0.20  &  86.80 $\pm$ 0.20 \\
    \hline
\textbf{Baseline-Cosine}  & \emph{ViT-S}  & 57.75 $\pm$ 0.16 & 72.15 $\pm$ 0.12   \\ 
\textbf{Ours-Cosine}  & \emph{ViT-S}  & {\color{red}78.89  $\pm$ 0.18} & 87.73  $\pm$ 0.11   \\ 
\textbf{Ours-Classifier}  & \emph{ViT-S}  & {78.88}  $\pm$ 0.18 &{\color{red}90.50 $\pm$ 0.09}\\ 

    \bottomrule
  \end{tabular}}
 
  \caption{Comparison  with  the state-of-the-art  5-way 1-shot and 5-way 5-shot performance  with  95$\%$  confidence  intervals  on \textbf{CIFAR-FS}. Top three results are shown in {\color{red}red}, {\color{blue}blue} and {\color{green}green}}.
  \label{tab:sota2}
\end{table}

\begin{table}[h]
\centering
 \normalsize 
 \scalebox{0.83}{
  \begin{tabular}{m{3.2cm}<{\centering} c cc}
    
    \toprule
    \multicolumn{1}{c }{\multirow{2}{*}{\textbf{Method}}} & 
    \multicolumn{1}{c }{\multirow{2}{*}{\textbf{backbone}}} & \multicolumn{2}{c}{\textbf{FC100}}  \\
    \multicolumn{1}{c }{}  &  \multicolumn{1}{c }{} & \multicolumn{1}{c}{\textbf{1-shot}}  & \multicolumn{1}{c}{\textbf{5-shot}}  \\
    \midrule
    {DeepEMD}\cite{DeepEMD} &    \emph{ResNet-12} &   46.47  $\pm$ 0.78  &   63.22  $\pm$ 0.71\\
    {IE}\cite{IE}  &  \emph{ResNet-12} &   {\color{blue}47.76   $\pm$ 0.77}  &   {\color{blue}65.30 $\pm$ 0.76}\\
    {BML}\cite{BML}  &  \emph{ResNet-12} &   45.00  $\pm$ 0.41  &  63.03   $\pm$ 0.41\\
    {ALFA+MeTAL}\cite{ALFA+MeTAL}  &  \emph{ResNet-12} &   44.54   $\pm$ 0.50  &  58.44 $\pm$ 0.42 \\
    {MixtFSL}\cite{MixtFSL}  &  \emph{ResNet-12} &   41.50   $\pm$ 0.67  &  58.39 $\pm$ 0.62 \\
    {PAL}\cite{PAL}  &  \emph{ResNet-12} &   {\color{green}47.20   $\pm$ 0.60}  &  {\color{green}64.00 $\pm$ 0.60} \\
    {TPMN}\cite{TPMN}  &  \emph{ResNet-12} &   46.93   $\pm$ 0.71  &  63.26 $\pm$ 0.74 \\
    {MN +  MC}\cite{MetaNavigator}  &  \emph{ResNet-12} &   46.40   $\pm$ 0.81  &  61.33 $\pm$ 0.71 \\
    {ConstellationNet}\cite{ConstellationNet}  &  \emph{ResNet-12} &   43.80   $\pm$ 0.20  &  59.70 $\pm$ 0.20 \\
    \hline
    \textbf{Baseline-Cosine}  & \emph{ViT-S}  & 40.83 $\pm$ 0.15 & 50.93 $\pm$ 0.15 \\ 
    \textbf{Ours-Cosine}  & \emph{ViT-S}  & {\color{red}48.27 $\pm$ 0.15} & 61.49 $\pm$ 0.15   \\ 
    \textbf{Ours-Classifier}  & \emph{ViT-S}  & 48.15 $\pm$ 0.16 & {\color{red}66.42 $\pm$ 0.16}\\ 
    \bottomrule
  \end{tabular}}
  \caption{Comparison  with  the  state-of-the-art  5-way 1-shot and 5-way 5-shot performance  with  95$\%$  confidence  intervals  on \textbf{FC100}. Top three results are shown in {\color{red}red}, {\color{blue}blue} and {\color{green}green}.}
  \label{tab:sota3}
\end{table}

\noindent{\textbf{Learnable Parameters Comparision.}} In Table~\ref{tab:sota_para}, we can find that more learnable parameters don't lead to better performance directly. The backbone of {METAQDA}$_{\;\rm{\,ICCV21}}$ has much more parameters than IE$_{\;\rm{\,CVPR21}}$ and HCTransformers, but get inferior performances. Compared with the ViT-S backbone and DINO$_{\;\rm{\,ICCV21}}$, HCTransformers get impressive improvements. It indicates that the proposed attribute surrogates learning and spectral tokens pooling are very important to utlize the strong learning abilities of transformers. Although we have more parameters than IE$_{\;\rm{\,CVPR21}}$ with the ResNet-12 backbone, we argue that our contribution is improving the data efficiency for transformers, and thus make them suitable for few-shot learning.

\begin{table}[h]
	\centering
	\scalebox{0.7}{
		\begin{tabular}{c c c c c}
			\toprule
			\textbf{Method} & \textbf{Backbone} & \textbf{Params} & \textbf{1-shot} & \textbf{5-shot}\\
			\midrule
			IE$_{\;\rm{\,CVPR21}}$ & ResNet-12 & 12.4M & 69.28$_{\pm 0.80}$ & 85.16$_{\pm 0.52}$ \\
			{METAQDA}$_{\;\rm{\,ICCV21}}$      & WRN-28-10 & 36.5M & 67.38$_{\pm 0.55}$ & 84.27$_{\pm 0.75}$ \\ 
			Cosine Distance                                  & ResNet-50 & 23M   & 59.28$_{\pm 0.20}$ & 72.68$_{\pm 0.16}$ \\
			Cosine Distance                                  & ViT-S     & 21M   & 52.92$_{\pm 0.17}$ & 65.04$_{\pm 0.14}$ \\
			DINO$_{\;\rm{\,ICCV21}}$({\color{blue}baseline}) & ViT-S     & 21M   & {\color{blue}61.57$_{\pm 0.16}$} & {\color{blue}75.51$_{\pm 0.12}$} \\ \hline
			HCTransformers 1                               & ViT-S     & 21M   & 71.27$_{\pm 0.17}$ & \textbf{85.66$_{\pm 0.10}$} \\
			HCTransformers 2                             & ViT-S     & 21M   & {\color{red}\textbf{74.74$_{\pm 0.17}$}} & {\color{red}\textbf{89.19$_{\pm 0.13}$}} \\
			HCTransformers 3                               & ViT-S     & 21M   & \textbf{72.66$_{\pm 0.17}$} & \textbf{85.66$_{\pm 0.10}$} \\
			\bottomrule
	\end{tabular}}
	\caption{Comparison of state-of-the-art algorithms with different backbones on {\em mini}Imagenet.}
	\label{tab:sota_para}
\end{table}

\noindent{\textbf{Computational Cost of Spectral Tokens Pooling.}} In Table~\ref{tab:time}, we list the training time cost of different modules in HCTransformers. It can be find that the spectral tokens pooling are relatively slow in our whole pipeline. But when compared with the first training stage, the time spent is still affordable because it needs only several epochs to train the second and third sets of transformers. 

\begin{table}[]
    \centering
    \scalebox{0.82}{
	\begin{tabular}{c|c|clc|cc}
		\hline
		\multirow{2}{*}{} & Stage 1(400 epoch) & \multicolumn{3}{c|}{Stage 2(2 epoch)}     & \multicolumn{2}{c}{Stage 3(2 epoch)} \\ \cline{2-7} 
		& ViT               & \multicolumn{2}{c}{Pooling} & ViT          & Pooling             & ViT             \\ \hline
		Time              & 21.1h             & \multicolumn{2}{c}{0.33 h}     & 0.25 h          & 0.12 h                 & 0.09 h             \\ \hline
	\end{tabular}}
	\caption{The amount of training time spent at each stage on 8 Nvidia RTX 3090 GPUs on {\em mini}ImageNet. }
\label{tab:time}
\end{table}

\subsection{Ablation Studies}
To investigate the contributions of different components, we conduct thorough experiments in this section. In the following, we use the cosine classifier for query labels prediction.\\

\begin{table}[htbp]
  \centering
  \scalebox{0.95}{
  \begin{tabular}{c c c c}
    \toprule
    \textbf{Method} & \textbf{Loss}  & \textbf{1-shot} & \textbf{5-shot} \\
    \midrule
    DINO &  - & $61.57 \pm 0.16$ & 75.51  $\pm$ 0.12\\
    DINO & CE &  66.81 $\pm$ 0.17 & 80.27 $\pm$ 0.12 \\
    DINO & PTH &  63.17  $\pm$ 0.16 & 78.59  $\pm$ 0.12\\
    DINO & CLS &  68.95  $\pm$ 0.17 & 82.83  $\pm$ 0.11\\
    
    \hline
    \textbf{Ours} & CLS+PTH &  \textbf{71.27  $\pm$ 0.17} & \textbf{84.68  $\pm$ 0.10}\\
    \bottomrule
  \end{tabular}}
  \begin{tablenotes} 
  \centering
		\item \small{CE: cross-entropy loss, PTH: pth loss,  CLS: cls loss.}
  \end{tablenotes}
  \caption{The results of the first student transformer trained with different supervision on \textit{ mini}Imagenet. All models are based on the DINO baseline. "CE" stands for the combination of a cross-entropy loss (as that in ViT~\cite{dosovitskiy2020image}) and the DINO loss. "PTH" stands for the combination of the patch surrogate loss and the DINO loss. "CLS" stands for the combination of the class surrogate loss and the DINO loss. "CLS+PTH" stands for the full combination of the class surrogate loss, the patch surrogate loss and the DINO loss.}
  \label{tab:losses}
\end{table}
\begin{table}[htb]
  \centering
  \begin{tabular}{c c c}
    \toprule
    \bm{$\beta$} &  \textbf{1-shot} & \textbf{5-shot} \\
    \midrule
    1 & 61.45  $\pm$ 0.16 & 78.59  $\pm$ 0.12\\
    \textbf{0.1} & \textbf{71.27  $\pm$ 0.17} & \textbf{84.68  $\pm$ 0.10}\\
    0.01 & 70.40  $\pm$ 0.16 & 84.07  $\pm$ 0.10\\
    \bottomrule
  \end{tabular}
  \caption{Test of the choice of different $\beta$ about the patch surrogate loss in the first student transformer on \textit{ mini}Imagenet.}
  \label{tab:beta}
\end{table}
\begin{figure*}[h]  
\centering  
\includegraphics[width=0.99\linewidth]{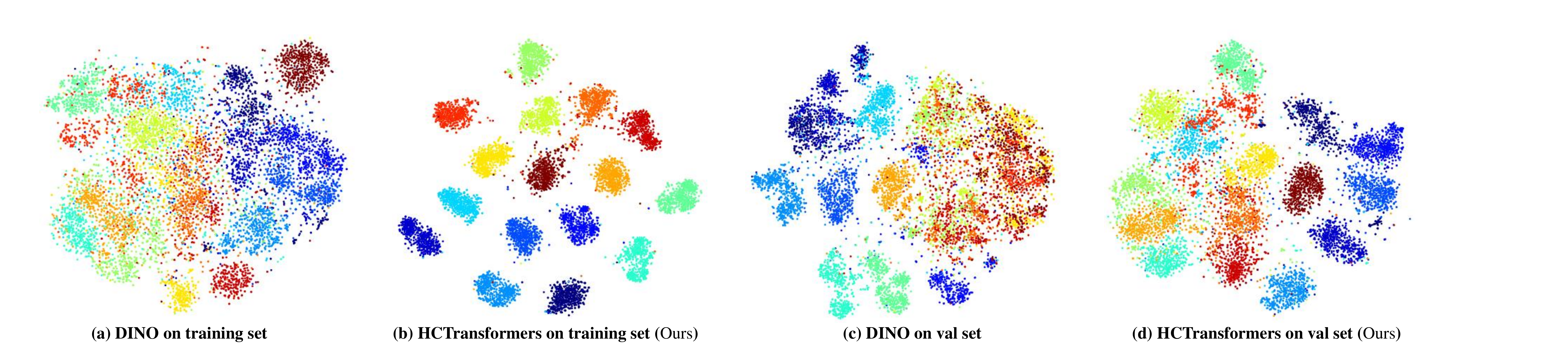}  
\caption{Visualizing features of DINO and HTransformers by t-SNE~\cite{van2008visualizing}   on both train and validation sets on \emph{mini}Imagenet. Points with the same color correspond to the same category. (a) and (b) show that when equipped with semantic surrogates, more distinctive features can be learned compared with DINO baseline. Feature distribution patterns on validation set(i.e., (c) and (d)) illustrate that our method can generalize well on unseen data.}  
\label{fig:tsne}
\end{figure*}

\noindent\textbf{Whether the latent supervision propagation is helpful?} To demonstrate the effectiveness of our proposed latent supervision propagation scheme, we conduct a series of experiments with different settings on \emph{mini}Imagenet in the first training stage. Results in \cref{tab:losses} show that our proposed scheme greatly improves over DINO baseline by 9.70\% for 1-shot setting and 9.17\% for 5-shot setting. 
To explore if the patch and class surrogate losses bring in benefits, we replace them with the commonly used cross entropy loss as in ViT to supervise the parameter learning together with the DINO loss. The 5-way 1-shot and 5-way 5-shot performances drop by 4.46\% and 4.41\% respectively. It validates our assumption that when there is few labeled data, supervising transformers with one-hot vectors will show inferior generalization ability. 

We also test the effects of the class and patch surrogate losses individually. When we remove the patch token loss for Eq.~\ref{eq:final_1_loss}, the 5-way 1-shot and 5-way 5-shot accuracy drop by 2.3\% and 1.85\%. If we remove the class token loss from Eq.~\ref{eq:final_1_loss}, the performances drop to 63.17\% and 78.59\% which indicates that the class surrogate loss is key to get good performance. All these experiments show that the two surrogate losses are effective in training transformers.   

We try different weights $\beta$ for the patch surrogate loss by setting it to 1, 0.1, 0.01, respectively. \cref{tab:beta} shows that setting $\beta$ to 0.1 is most suitable for the current setting. \\

\begin{table}[h]
  \centering
  \begin{tabular}{c c c}
    \toprule
    \textbf{resolution/patch} & \textbf{1-shot} & \textbf{5-shot} \\
    \midrule
    $56^2/2$ & 71.99  $\pm$ 0.17 & 85.04  $\pm$ 0.11\\
    $112^2/4$ & 72.80  $\pm$ 0.17 & 85.93  $\pm$ 0.11\\
    \bm{$224^2/8$} & \textbf{73.28  $\pm$ 0.16} & \textbf{86.49  $\pm$ 0.11}\\
    
    \bottomrule
  \end{tabular}
  \caption{The performances of the first student transformer which takes in images with different resolutions/patch sizes on CIFAR-FS.}
  \label{tab:resolution}
\end{table}
\begin{table}[h]
  \centering
  \begin{tabular}{c c c}
    \toprule
    \textbf{model} & \bm{$84^2$} & \bm{$224^2$} \\
    \midrule
    IE~\cite{IE} & 66.82  $\pm$ 0.80 & 60.88  $\pm$ 0.81\\
    DeepEMD~\cite{DeepEMD} & 65.91 $\pm$ 0.82 & 63.07 $\pm$ 0.81\\
    \bottomrule
  \end{tabular}
  \caption{The influence of image resolutions on two state-of-the-art methods on \emph{mini}Imagenet for 5-way 1-shot classification.}
  \label{tab:resolution1}
\end{table}

\noindent \textbf{How is the input resolution affects the results?} In our implementation, images are resized into $224 \times 224$, which is higher than traditional CNN-based methods, such as IE~\cite{IE,S2M2}. Multiple experiments are conducted to prove such a resolution yields is suitable for ViT~\cite{dosovitskiy2020image}, while other state-of-the-art methods including IE~\cite{IE} and DeepEMD~\cite{DeepEMD}, fail to utilize high-resolution inputs. To test the performance of our HCTransformers when taking in images with different resolutions, we conduct experiments on CIFAR-FS~\cite{CIFAR-FS}. We resize images into $32 \times 32$ to $54 \times 54$, $112 \times 112$, and $224 \times 224$. The patch sizes are then $2 \times 2$,$4 \times 4$ and $8 \times 8$ respectively. Results of different settings are listed in \cref{tab:resolution}, showing that higher resolutions lead to better performance. We think that a higher resolution will make transformers get more information in the input patches and produce stable patch representations. 

We also run other state-of-the-art methods on \emph{mini}Imagenet in 5-way 1-shot setting by upgrading the resolution to $224 \times 224$, and list experimental results in \cref{tab:resolution1}. Their performances with high resolution are worse than their original ones. Results suggest that higher resolution will always lead to performance improvement in few-shot learning tasks. Similar to ours, BML~\cite{BML} also made experiments on DeepEMD with high-resolution inputs but failed to obtain satisfactory results.\\

\noindent \textbf{Is the end-to-end training necessary in the second training stage?}  To test whether it will be better to train the last two sets of transformers one-by-one, we train the second set of transformers and then freeze their parameters to train the third set of transformers. As shown in \cref{tab:pooling}, training the last two sets of transformers one-by-one leads to comparable but slightly lower performance than the end-to-end one.  The reason may be that jointly training the latter two sets of transformers end-to-end may help the second set of transformers to learn better features.
\begin{table}[t]
  \centering
  \scalebox{0.87}{
  \begin{tabular}{c c c c}
    \toprule
     \textbf{training mode} & \textbf{stage1} & \textbf{stage2} & \textbf{stage3} \\
    \midrule
    
    one-by-one & 71.27  $\pm$ 0.17 & 74.40  $\pm$ 0.17  & 73.01 $\pm$ 0.18\\
    end-to-end & 71.27  $\pm$ 0.17 & \textbf{74.74  $\pm$ 0.17}  & 72.66 $\pm$ 0.18\\
    \bottomrule
  \end{tabular}
  }
  \caption{Ablations of training the last two sets of transformers in a one-by-one or end-to-end manner.}
  \label{tab:pooling}
\end{table}

\section{Limitations}
HCTransformers achieve good results in few-shot classification, but the current setting requires images to have a rather large resolution to pacify possible chaos within the patch to construct a stable patch-level representation. This may lead to unsatisfactory performance when the input images are in low resolutions. Besides, spectral tokens pooling is time-consuming. It will limit the usage of HCTransformers in many real applications.  

\section{Conclusion}
We propose hierarchically cascaded transformers that can improve the data efficiency to tackle the task of few-shot image classification. Despite vision transformer's data-hungry nature, we achieved good performance in few-shot learning tasks. Our proposed method introduces a latent supervision propagation technique that implicitly supervises the parameter learning with attribute surrogates that can be learnt. We propose a scheme to integrate patch tokens that can work in complementary with $[cls]$ token. Also, spectral tokens pooling is proposed to embed the object/scene layout and semantic relationship among tokens for transformers. Our proposed HCTransformers not only outperform the DINO baseline significantly, but also surpass previous state-of-the-art methods by clear margins on \emph{mini}Imagenet, \emph{tiered}Imagenet datasets, CIFAR-FS and FC100. 

\section{Acknowledgement}
This work was partly supported by the National Natural Science Foundation of China under Grant Nos. 62106051 and the Shanghai Pujiang Program Nos. 21PJ1400600.

{\small
\bibliographystyle{ieee_fullname}
\bibliography{egbib}
}
\clearpage
\input{supp_updated}

\end{document}

%% file: supp_updated.tex
\makeatletter
\def\@thanks{}
\makeatother

\title{\emph{Supplementary Material} for \\
Attribute Surrogates Learning and Spectral Tokens Pooling
in Transformers for Few-shot Learning}

\makeatletter
\renewcommand{\@maketitle}{
\newpage
 \null
 \vskip 2em%
 \begin{center}%
  {\LARGE \@title \par}%
 \end{center}%
 \par} \makeatother

\renewcommand\thesection{\Alph{section}}
\setcounter{section}{0}
\setcounter{table}{0}
\setcounter{figure}{0}

    \onecolumn
    \maketitle
    \setlength{\parindent}{1em} 
    \setlength{\parskip}{0em}
    

\section{More Ablation Results}
Extended experiments are conducted to further validate our design's effectiveness. 
\paragraph{Do class and patch surrogates need different semantic spaces?}In our proposed methods, we expect class surrogates and patch surrogates to reside in different semantic spaces to complete each other by solely feeding class output into a MLP layer. Experimental results show that without this re-projecting procedure, performance drops significantly by 29.5\%. It suggests that when constrained within the same feature space, supervision on class surrogate and patch surrogate might lead to severe confusion that hinders efficient learning.  

\paragraph{Are local views harmful for surrogate learning?} For every input image, our training framework produces several local and global views for self-supervision. When imposing additional surrogate-level supervision, we only attend to these global views for their better information perseverance. Experimental results show that with both global and local views supervised, performance drops by 1.9\% , corresponding to the second line with views global+local in Table \ref{tab:mlp}. It suggests that local views tend to introduce noise for surrogate learning, which is natural considering the non-neglected object truncation issues in local views.

\paragraph{Why not use patch loss in latter stages?} 
In our design, we choose to let latter transformers inherit $[class]$ token from previous stage for initialization and impose no surrogate supervision on patch levels. We conduct experiments where patch-level supervision is preserved in latter transformer sets. As shown in \cref{tab:pooling2}, with patch-level supervision, results barely improve over the basis of stage 1 and are inferior to our design with class-level supervision only. We suggest that the reason lies in that at latter stages, complementary characteristics of $[cls]$ surrogates is key to further performance improvement.  With patch loss added, network will lose its focus in $[cls]$ surrogate learning and the intertwined training will lose previous $[cls]$ surrogates' good properties.
\paragraph{Do higher resolutions always lead to better performance?}
Our proposed method use image with resolution $224\times 224$ as input, which is different from existing SOTA methods. To prove that our performance elevation is not a resolution gain, we conduct experiments with two SOTA methods in $224\times 224$ settings in Table 7 of our paper. Results show that higher resolutions lead to performance drop because of the increased tendency to overfitting for these methods. For supplementary, results with Meta-baseline \cite{chen2021meta} are listed in \ref{tab:sota1}. These results show that our method's improvement is not introduced by higher resolution and possess stronger generalization ability.  
\section{Qualitative Results}
\begin{figure*}[h]
	\centering
	\includegraphics[width=1\linewidth]{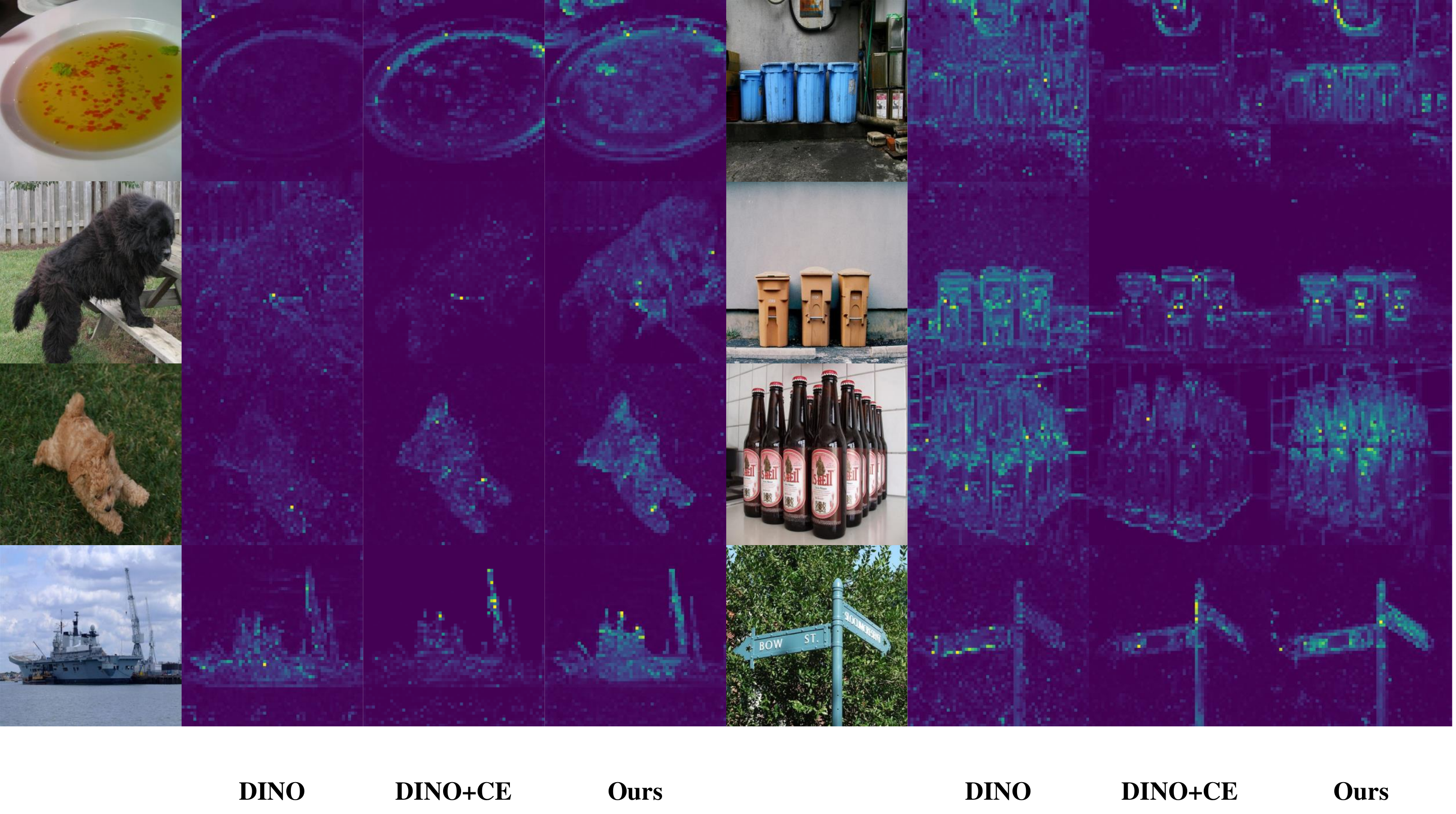}
	\caption{Visualization of Self-attention maps with the $[cls]$ token in final layers by DINO baseline, DINO + CE(with class-level supervision) and our proposed method. }
	\label{Fig:1}
\end{figure*}

\begin{figure*}[h]
	\centering
	\includegraphics[width=0.75\linewidth]{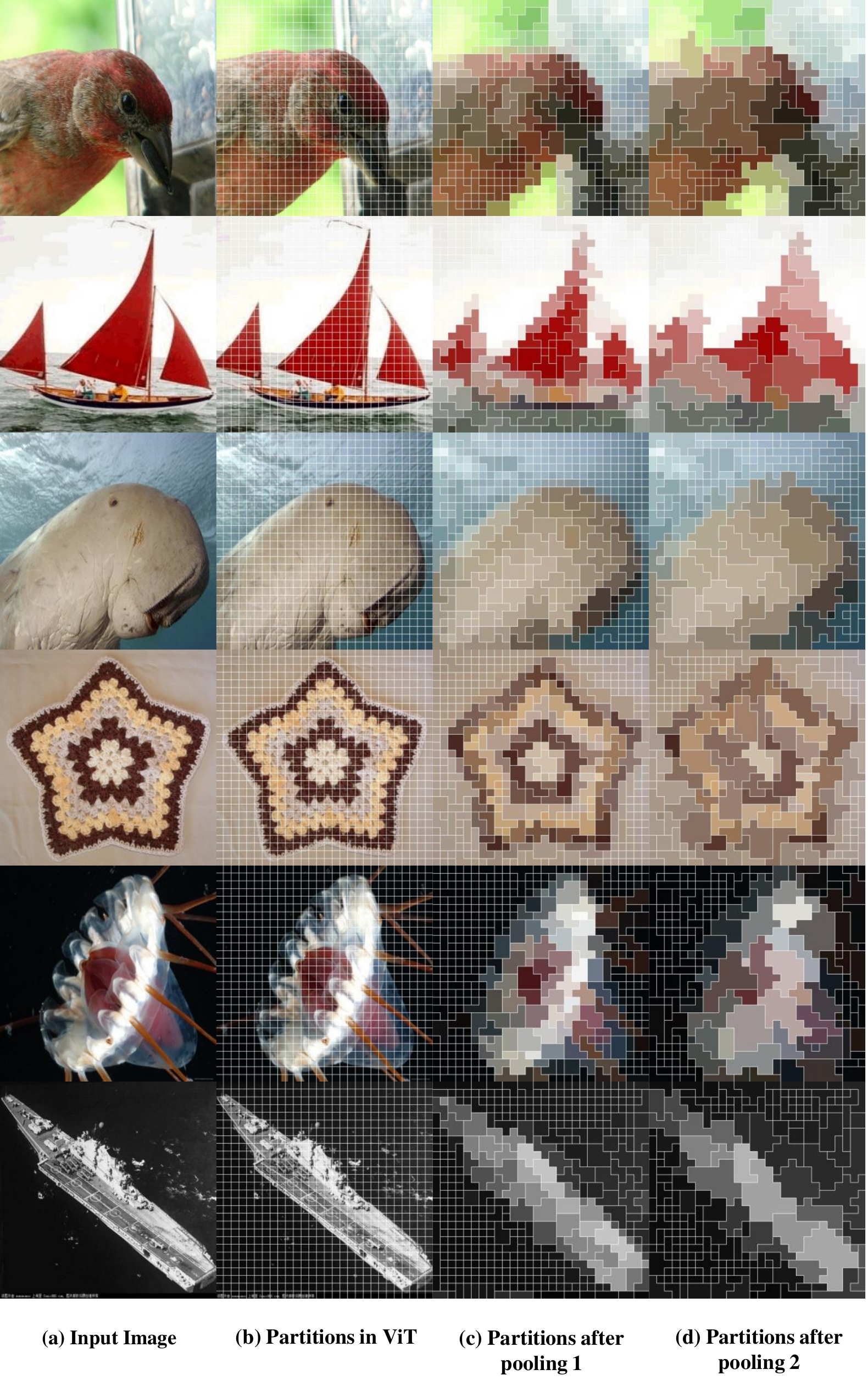}
	\caption{More visualization of tokens pooling process. After spectral tokens pooling operations, adjacent tokens with similar semantics
are clustered into one. (c) and (d) shows that our clustering results
are well consistent with the image’s basic structure. The pixel colors in the same cluster are averaged.}
	\label{Fig:2}
\end{figure*}

To further validate that our method can retrieve meaningful semantic representations with small datasets, we visualize self-attention maps associated with $[cls]$ token and show results from our retrained DINO baseline and DINO +CE(with class-level cross entropy supervision) in \cref{Fig:1} for fair comparison. Results show that our method focuses more on foreground information than the other two methods. In addition, we listed more visual results of two consecutive spectral tokens pooling procedure in \cref{Fig:2}.
\begin{table*}[t]
  \centering
  
  \begin{tabular}{c c c c c c}
    \toprule
    \textbf{Method}&$\textbf f_c\textbf{(x)}$ &\textbf{P(x)} &\textbf{views} & \textbf{1-shot} & \textbf{5-shot} \\
    \midrule
    DINO & $\checkmark$ & -& global& 41.79 $\pm$ 0.17 & 56.27 $\pm$ 0.15\\
    DINO &  - & \checkmark & global+local& 69.37 $\pm$ 0.16  & 82.99 $\pm$ 0.11\\
    
    \hline
    \textbf{Ours} &  - & \checkmark & global&  \textbf{71.27  $\pm$ 0.17} & \textbf{84.68  $\pm$ 0.10}\\
    \bottomrule
  \end{tabular}
  \begin{tablenotes} 
  \centering
		
		\item \small{$f_c(x)$: the encoded $[cls]$ token,
		$P(x)$: the $[cls]$ token after projection head}
  \end{tablenotes}
  \caption{Results of the first student transformer trained with different surrogate space and image view settings on \textit{ mini}Imagenet. All models are based on the DINO baseline.  Two choices for class surrogate are tested : $f_c(x)$, the $[cls]$ token extracted from ViT encoder, and $P(x)$, projection of $[cls]$ token via classification head. To explore impacts of local views for surrogate-level supervision, experiment with all views supervised is conducted for fair comparison.}
  \label{tab:mlp}

\end{table*}
\begin{table*}[htbp]
  \centering
  
  \begin{tabular}{ c c c c}
    \toprule
    \textbf{Loss} & \textbf{stage1} & \textbf{stage2} & \textbf{stage3} \\
    \midrule
    DINO+CLS+PTH & 71.27 $\pm$ 0.17 &  71.25 $\pm$ 0.17 & 71.19 $\pm$ 0.17\\
    \hline
    DINO+CLS & -  & \textbf{74.74 $\pm$ 0.17}  & 72.66 $\pm$ 0.18\\
    
    \bottomrule
  \end{tabular}
  \begin{tablenotes} 
  \centering
		\item \small{DINO+CLS: combination of the class surrogate loss and the DINO loss.}
		\vspace{0.5pt}
		\item \small{DINO+CLS+PTH: full combination of the class surrogate loss, the patch surrogate loss and the DINO loss.}
  \end{tablenotes}
  \caption{Results of supervising learning process for the latter two transformer sets with or without patch surrogates.}
  
 \label{tab:pooling2}
\end{table*}
\begin{table*}[h]
\centering

  \small
  \begin{tabular}{c c cc cc}
    
    \toprule
    \multicolumn{1}{c}{\multirow{2}{*}{\textbf{Method}}}  & 
    \multicolumn{1}{c}{\multirow{2}{*}{\textbf{resolution}}} & \multicolumn{2}{c}{\textbf{\textit{mini}Imagenet}} & \multicolumn{2}{c}{\textbf{\textit{tiered}Imagenet}} \\
    \multicolumn{1}{c}{}   & \multicolumn{1}{c}{} & \textbf{1-shot}  & \multicolumn{1}{c}{\textbf{5-shot}} & \textbf{1-shot}  & \textbf{5-shot} \\
    \midrule

{Meta-baseline\cite{chen2021meta}} &    \emph{$84^2$} &   63.17  $\pm$ 0.23  &   79.26  $\pm$ 0.17  &   68.62  $\pm$ 0.27  &   83.74  $\pm$ 0.18 \\
{Meta-baseline\cite{chen2021meta}} &    \emph{$224^2$} &   67.20  $\pm$ 0.23  &   81.19  $\pm$ 0.16  &   70.28  $\pm$ 0.27  &   82.24  $\pm$ 0.20 \\

\hline

\textbf{Ours-Cosine}  & \emph{$224^2$}  & 74.74 $\pm$ 0.17 & 85.66  $\pm$ 0.10  & 79.67 $\pm$ 0.20  & 89.27  $\pm$ 0.13 \\ 

    \bottomrule
  \end{tabular}
 
  \caption{Comparison with the state-of-the-art 5-way 1-shot and 5-way 5-shot performance with 95\% confidence intervals on different resolutions on \textbf{\textit{mini}Imagenet} and \textbf{\textit{tiered}Imagenet}.}
  \label{tab:sota1}
\end{table*}


\FloatBarrier